\documentclass{article}

\usepackage[preprint, nonatbib]{neurips_2022}

\usepackage[utf8]{inputenc} 
\usepackage[T1]{fontenc}    
\usepackage{hyperref}       
\usepackage{url}            
\usepackage{booktabs}       
\usepackage{amsfonts}       
\usepackage{nicefrac}       
\usepackage{microtype}      
\usepackage[dvipsnames]{xcolor}         
\usepackage{graphicx}
\usepackage{multirow}
\usepackage{amsmath}

\newcommand{\mbf}{\mathbf}

\usepackage{xspace}
\makeatletter
\DeclareRobustCommand\onedot{\futurelet\@let@token\@onedot}
\def\@onedot{\ifx\@let@token.\else.\null\fi\xspace}

\def\eg{\emph{e.g}\onedot}

\def\etal{\emph{et al}\onedot}
\makeatother

\title{HyperTime: Implicit Neural Representation\\for Time Series}

\author{%
  Elizabeth Fons\\
  J. P. Morgan AI Research\\
   \And
   Alejandro Sztrajman \\
   University College London \\
   \And
   Yousef El-laham \\
   J. P. Morgan AI Research \\
   \And
   Alexandros Iosifidis \\
   Aarhus University \\
   \And
   Svitlana Vyetrenko \\
   J. P. Morgan AI Research \\
}

\begin{document}

\maketitle

\begin{abstract}
Implicit neural representations (INRs) have recently emerged as a powerful tool that provides an accurate and resolution-independent encoding of data. Their robustness as general approximators has been shown in a wide variety of data sources, with applications on image, sound, and 3D scene representation. However, little attention has been given to leveraging these architectures for the representation and analysis of time series data. In this paper, we analyze the representation of time series using INRs, 
comparing different activation functions in terms of reconstruction accuracy and training convergence speed. We show how these networks can be leveraged for the imputation of time series, with applications on both univariate and multivariate data. Finally, we propose a hypernetwork architecture that leverages INRs to learn a compressed latent representation of an entire time series dataset. We introduce an FFT-based loss to guide training so that all frequencies are preserved in the time series.
We show that this network can be used to encode time series as INRs, and their embeddings can be interpolated to generate new time series from existing ones. We evaluate our generative method by using it for data augmentation, and show that it is competitive against current state-of-the-art approaches for augmentation of time series.
\end{abstract}

\section{Introduction}
\label{sec:intro}

Modeling time series data has been a key topic of research for many years, constituting a crucial component of applications in a wide variety of areas such as climate modeling, medicine, biology, retail and finance~\cite{lim2021}.
Traditional methods for time series modeling have relied on parametric models informed by expert knowledge. However, the development of modern machine learning methods has provided purely data-driven techniques to learn temporal relationships. In particular, neural network-based methods have gained popularity in recent times, with applications on a wide range of tasks, such as time series classification~\cite{Fawaz2020}, clustering~\cite{Ma2019, deep_clustering}, segmentation~\cite{Perslev2019, zeng2022segtime}, anomaly detection~\cite{Choi2021, Xu2018, Hundman2018}, upsampling~\cite{Oh2020, Bellos2019ACN}, imputation~\cite{Liu2018, Luo2018, BRITS2018}, forecasting~\cite{lim2021, Torres2021DeepLF} and synthesis~\cite{alaa2021, timeGAN, pategan}. 
In particular, the generation of time series data for augmentation has remained as an open problem, and is currently gaining interest due to the large number of potential applications such as in medical and financial datasets, where data cannot be shared, either for privacy reasons or for proprietary restrictions~\cite{Jordon2021, pategan, Assefa2020, Coletta2021}.

In recent years, implicit neural representations (INRs) have gained popularity as an accurate and flexible method to parameterize signals, such as from image, video, audio and 3D scene data~\cite{sitzmann2020siren, mildenhall2020nerf}. Conventional methods for data encoding often rely on discrete representations, such as data grids, which are limited by their spatial resolution and present inherent discretization artifacts. In contrast, implicit neural representations encode data in terms of continuous functional relationships between signals, and thus are uncoupled to spatial resolution. In practical terms, INRs provide a new data representation framework that is resolution-independent, with many potential applications on time series data, where irregularly sampled and missing data are common occurrences~\cite{Fang2020survey}. However, there are currently no works exploring the suitability of INRs on time series representation and analysis.

In this work, we propose an implicit neural representation for univariate and multivariate time series data.
We compare the performance of different activation functions in terms of reconstruction accuracy and training convergence, and we formulate and compare different strategies for data imputation in time series, relying on INRs (Section~\ref{sec:exp:imputation}). Finally, we combine these representations with a hypernetwork architecture, in order to learn a prior over the space of time series. The training of our hypernetwork takes into account the accurate reconstruction of both the time series signals and their respective power spectra. This motivates us to propose a Fourier-based loss that proves to be crucial in guiding the learning process. The advantage of employing such a Fourier-based loss is that it allows our hypernetwork to preserve all frequencies in the time series representation.
In Section~\ref{sec:exp:generation}, we leverage the latent embeddings learned by the hypernetwork for the synthesis of new time series by interpolation, and show that our method performs competitively against recent state-of-the-art methods for time series augmentation.

\section{Related Work}
\label{sec:rw}

\paragraph{Implicit Neural Representations}

INRs 
(or coordinate-based neural networks) have recently gained popularity in computer vision applications. The usual implementation of INRs 
consists of a fully-connected neural network (MLP) that maps coordinates (\eg xyz-coordinates) to the corresponding values of the data, essentially encoding their functional relationship in the network. One of the main advantages of this approach for data representation, is that the information is encoded in a continuous/grid-free representation, that provides a built-in non-linear interpolation of the data. This avoids the usual artifacts that arise from discretization, and has been shown to combine flexible and accurate data representation, with high memory efficiency~\cite{sitzmann2020siren, tancik2020fourfeat}. Whilst INRs have been shown to work on data from diverse sources, such as video, images and audio~\cite{sitzmann2020siren, chen2021, shaham2021}, their recent popularity has been motivated by multiple applications in the representation of 3D scene data, such as 3D geometry~\cite{park2019deepsdf, mescheder2019, sitzmann2019metasdf, sitzmann2019srns} and object appearance~\cite{mildenhall2020nerf, sztrajman2021}.

In early architectures, INRs 
showed a lack of accuracy in the encoding of high-frequency details of signals. Mildenhall~\etal~\cite{mildenhall2020nerf} proposed positional encodings to address this issue, and Tancik~\etal~\cite{tancik2020fourfeat} further explored them, showing that by using Fourier-based features in the input layer, the network is able to learn the full spectrum of frequencies from data. Concurrently, Sitzmann~\etal~\cite{sitzmann2020siren} tackled the encoding of high-frequency data by proposing the use of sinusoidal activation functions (SIREN: Sinusoidal Representation Networks), and Benbarka~\etal~\cite{benbarka2022} showed the equivalence between Fourier features and single-layer SIRENs. Our INR architecture for time series data (Section~\ref{sec:form:repr}) is based on the SIREN architecture by Sitzmann~\etal. In Section~\ref{sec:exp:rec} we compare the performance of different activation layers, in terms of reconstruction accuracy and training convergence speed, for both univariate and multivariate time series.

\paragraph{Hypernetworks}
A hypernetwork is a neural network architecture designed to predict the weight values of a secondary neural network, denominated a HypoNetwork~\cite{sitzmann2019metasdf}. The concept of hypernetwork was formalized by Ha~\etal~\cite{ha2017}, drawing inspiration from methods in evolutionary computing~\cite{stanley2009}. Moreover, while convolutional encoders have been likened to the function of the human visual system~\cite{skorokhodov2021}, the analogy cannot be extended to convolutional decoders, and some authors have argued that hypernetworks much more closely match the behavior of the prefrontal cortex ~\cite{russin2020}.
Hypernetworks have been praised for their expressivity, compression due to weight sharing, and for their fast inference times\cite{skorokhodov2021}. They have been leveraged for multiple applications, including few-shot learning~\cite{rusu2018, zhao2020}, continual learning~\cite{Oswald2020Continual} and architecture search~\cite{zhang2018graph, brock2018smash}. Moreover, in the last two years some works have started to leverage hypernetworks for the training of INRs, 
enabling the learning of latent encodings of data, while also maintaining the flexible and accurate reconstruction of signals provided by INRs. This approach has been implemented with different hypernetwork architectures, to learn priors over image data~\cite{sitzmann2020siren, skorokhodov2021}, 3D scene geometry~\cite{littwin2019, sitzmann2019srns, sitzmann2019metasdf} and material appearance~\cite{sztrajman2021}. Tancik~\etal~\cite{tancik2020meta} leverage hypernetworks to speed-up the training of INRs by providing learned initializations of the network weights. Sitzmann~\etal~\cite{sitzmann2020siren} combine a set encoder with a hypernetwork decoder to learn a prior over INRs representing image data, and apply it for image in-painting. Our hypernetwork architecture from Section~\ref{sec:form} is similar to Sitzmann~\etal's, however we learn a prior over the space of time series and leverage it for new data synthesis through interpolation of the learned embeddings. Furthermore, our architecture implements a Fourier-based loss, which we show is crucial for the accurate reconstruction of time series datasets (Section~\ref{sec:exp:generation}).

\paragraph{Time Series Generation}

Realistic time series generation has been previously studied in the literature by using the generative adversarial networks (GANs). In TimeGAN architecture~\cite{Yoon2019TimeseriesGA}, realistic generation of temporal patterns was achieved by jointly optimizing with both supervised and adversarial objectives to learn an embedding space. QuantGAN~\cite{Wiese_2020} consists of a generator and discriminator functions represented by temporal convolutional networks, which allows it to synthesize long-range dependencies such as the presence of volatility clusters that are characteristic of financial time series. TimeVAE~\cite{desai2021timevae} was recently proposed as a variational autoencoder alternative to GAN-based time series generation. GANs and VAEs are typically used for creating statistical replicas of the training data, and not the distributionally new scenarios needed for data augmentation. More recently, Alaa~\etal~\cite{alaa2021} presented Fourier Flows, a normalizing flow model for time series data that leverages the frequency domain representation, currently considered together with TimeGAN as state-of-the-art for time series augmentation. 

Data augmentation is well established in computer vision tasks due to the simplicity of label-preserving geometric image transformation techniques, but it is still not widely used for time series with some early work being discussed in the literature~\cite{timeseries_augmentation}. For example, simple augmentation techniques applied to financial price time series such as adding noise or time warping were shown to improve the quality of next day price prediction model.~\cite{fons2021adaptive}

\section{Formulation}
\label{sec:form}

In this Section we describe the network architectures that we use to encode time series data (Subsection~\ref{sec:form:repr}), and the hypernetwork architecture (HyperTime) leveraged for prior learning and new data generation (Subsection~\ref{sec:form:gen}).

\subsection{Time Series Representation}
\label{sec:form:repr}

\begin{figure}[htb!]
    \centering
    \includegraphics[width=0.6\textwidth]{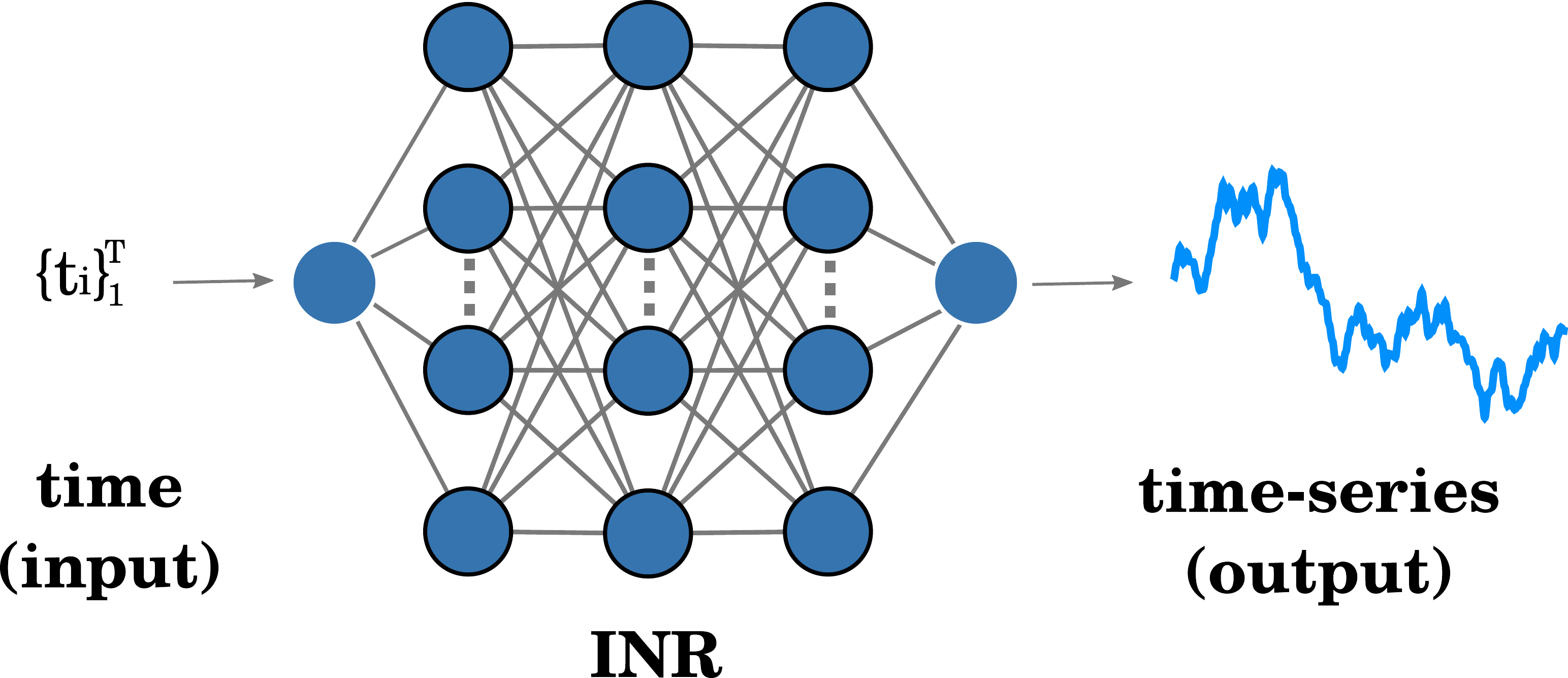}
    \caption{Diagram of the implicit neural representation (INR) for univariate time series. Neurons with a black border use sine activations. The network is composed of fully-connected layers with dimensions $1\times60\times60\times60\times1$.}
    \label{fig:inr}
\end{figure}

In Figure~\ref{fig:inr} we present a diagram of the INR 
used for univariate time series. The network is composed of fully-connected layers of dimensions $1\times60\times60\times60\times1$, with sine activations (SIREN~\cite{sitzmann2020siren}):
\begin{equation}
    \phi_i(\mbf{x}_i) = \text{sin}(\omega_0 \mbf{W}_i \mbf{x}_i + \mbf{b}_i)
    \label{eq:sine}
\end{equation}
where $\phi_i$ corresponds to the $i^{th}$ layer of the network. A general factor $\omega_0$ multiplying the network weights determines the order of magnitude of the frequencies that will be used to encode the signal.
Input and output of the INR are uni-dimensional, and correspond to the time coordinate $t$ and the time series evaluation $f(t)$. Training of the network is done in a supervised manner, with MSE loss, and takes less than $10 s$ in a GeForce GTX 1080 Ti GPU. After training, the network encodes a continuous representation of the functional relationship $f(t)$ for a single time series.

The architecture from Figure~\ref{fig:inr} can be modified to encode multivariate time series, by simply increasing the number of neurons of the output layer to match the number of channels of the signal. Due to weight-sharing, this adds a potential for data compression of the time series. In addition, the simultaneous encoding of multiple correlated channels can be leveraged for channel imputation, as we will show in Section~\ref{sec:exp:imputation}.

\subsection{Time Series Generation with HyperTime}
\label{sec:form:gen}

\begin{figure}[htb!]
    \centering
    \includegraphics[width=\textwidth]{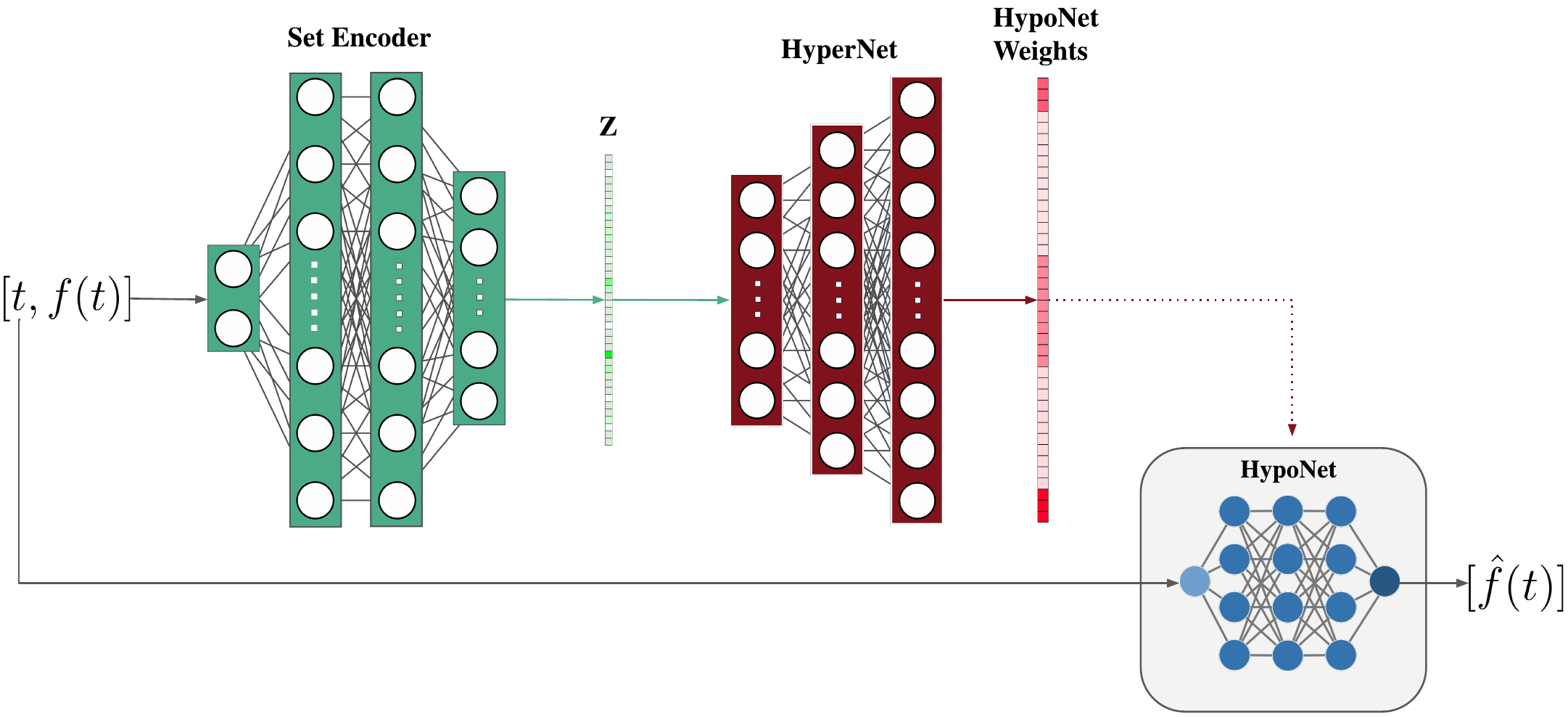}
    \caption{Diagram of HyperTime network architecture. Each pair of time-coordinate $t$ and time series $f(t)$ is encoded as a 40-values embedding $Z$ by the Set Encoder. The HyperNet decoder learns to predict HypoNet weights from the embeddings. During training, the output of the HyperNet is used to build a HypoNet and evaluate it on in the input time-coordinates. The loss is computed as a difference between $f(t)$ and the output of the HypoNet $\hat f(t)$.}
    \label{fig:hypertime}
\end{figure}

In Figure~\ref{fig:hypertime} we display a diagram of the HyperTime architecture, which allows us to leverage INRs to learn priors over the space of time series. The Set Encoder (green network), composed of SIREN layers~\cite{sitzmann2020siren} with dimensions $2\times128\times128\times40$, takes as input a pair of values, corresponding to the time-coordinate $t$ and the time series signal $f(t)$. Each pair of input values is thus encoded into a full $40$-values embedding and fed to the HyperNet decoder (red network), composed of fully-connected layers with ReLU activations (MLP), with dimensions $40\times128\times7500$. The output of the HyperNet is a one-dimensional $7500$-values embedding that contains the network weights of an INR which encodes the time series data from the input. The INR architecture used within HyperTime is the same described in the previous section, and illustrated in Figure~\ref{fig:inr}. Following previous works~\cite{sitzmann2019metasdf}, in order to avoid ambiguities we refer to these predicted INRs as HypoNets.

During the training of HyperTime, we use the weights predicted by the HyperNet decoder to instantiate a HypoNet and evaluate it on the input time-coordinate $t$, to produce the predicted time series value $\hat f(t)$. The entire chain of operations is implemented within the same differentiable pipeline, and hence the training loss can be computed as the difference between the ground truth time series signal $f(t)$ and the value predicted by the HypoNet $\hat f(t)$.

After the training of HyperTime, the Set Encoder is able to generate latent embeddings $Z$ for entire time series. In Section~\ref{sec:exp:generation}, we show that these embeddings can be interpolated to synthesize new time series signals from known ones, which can be leveraged for data augmentation (see additional material for a pseudo-code of the procedure).

\paragraph{Loss}

The training of HyperTime is done by optimizing the following loss, which contains an MSE reconstruction term $\mathcal{L}_\text{rec}$ and two regularization terms $\mathcal{L}_\text{weights}$ and $\mathcal{L}_\text{latent}$, for the network weights and the latent embeddings respectively:
\begin{equation}
\mathcal{L} = \underbrace{\frac{1}{N} \sum_{i=1}^N \left\| f(t_i) - \hat f(t_i) \right\|^2}_{\mathcal{L}_\text{rec}} +
              \lambda_1 \underbrace{\frac{1}{W} \sum_{j=1}^W w^2_j}_{\mathcal{L}_\text{weights}} +
              \lambda_2 \underbrace{\frac{1}{Z} \sum_{k=1}^Z z^2_k}_{\mathcal{L}_\text{latent}} +
              \lambda_3 \mathcal{L}_\text{FFT}
\label{eq:loss_hypertime}
\end{equation}

In addition, we introduce a Fourier-based loss $\mathcal{L}_\text{FFT}$ that focuses on the accurate reconstruction of the power spectrum of the ground truth signal (see Supplement for more details):
\begin{equation}
    \mathcal{L}_\text{FFT} = \frac{1}{N} \sum_{i=1}^N \left\|\text{FFT}[f(t)]_i - \text{FFT}[\hat f(t)]_i \right\|.
    \label{eq:fft_loss}
\end{equation}
In Section~\ref{sec:exp:generation}, we show that $\mathcal{L}_\text{FFT}$ is crucial for the accurate reconstruction of the time series signals.

\section{Experiments}
\label{sec:exp}

\subsection{Reconstruction}
\label{sec:exp:rec}

\begin{figure}[htb!]
    \centering
    {\scriptsize \hspace{4em}Losses\hspace{22em}Reconstructions\par\medskip}
    {\tiny \raisebox{8em}{\rotatebox[origin=c]{90}{MSE (log scale)}}}
    \includegraphics[width=0.4\textwidth, height=10em, trim={20px 20px 0 20px},clip]{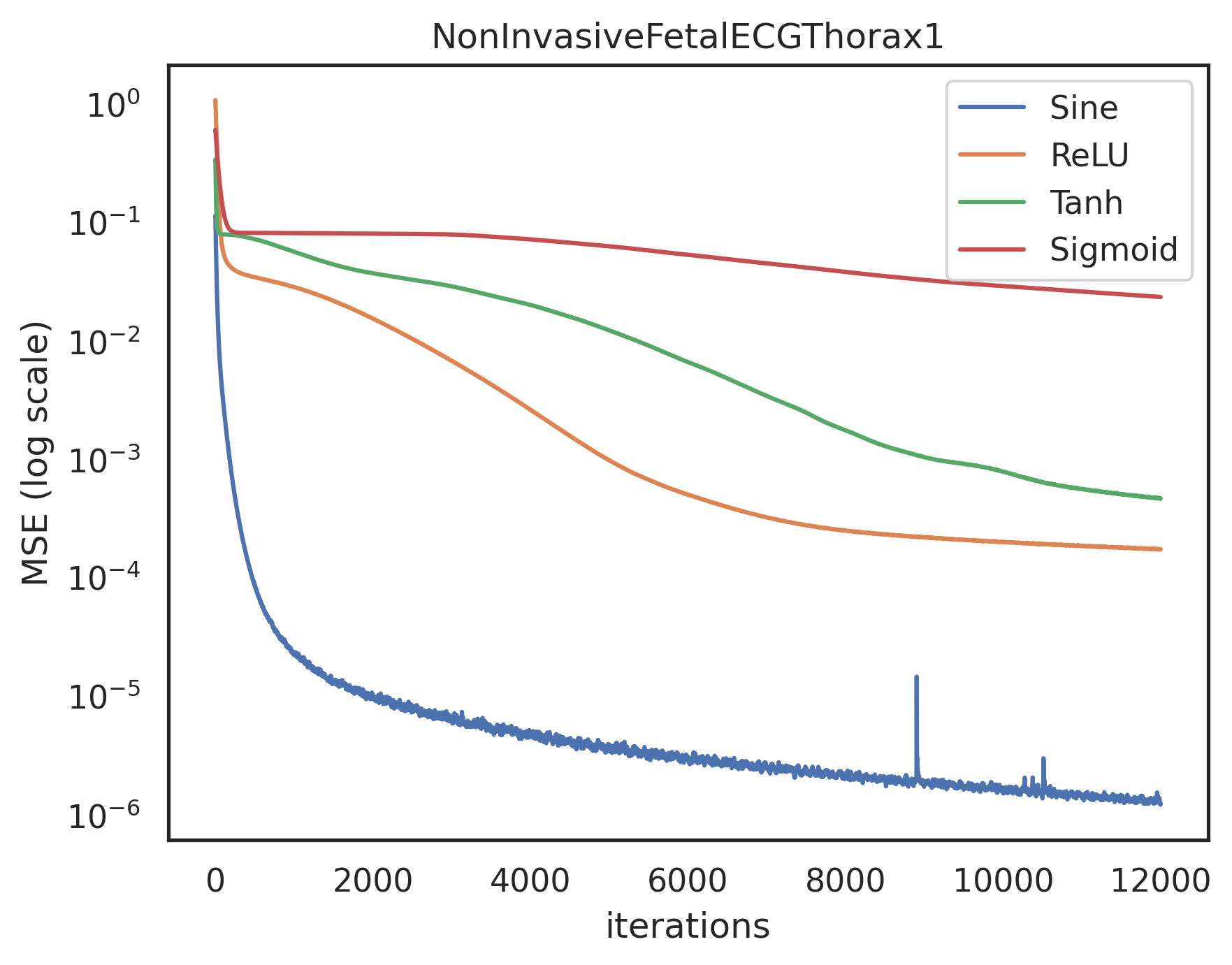} \hspace{2em}
    {\tiny \raisebox{8em}{\rotatebox[origin=c]{90}{Time-series value}}}
    \includegraphics[width=0.4\textwidth, height=10em, trim={20px 30px 0 0px},clip]{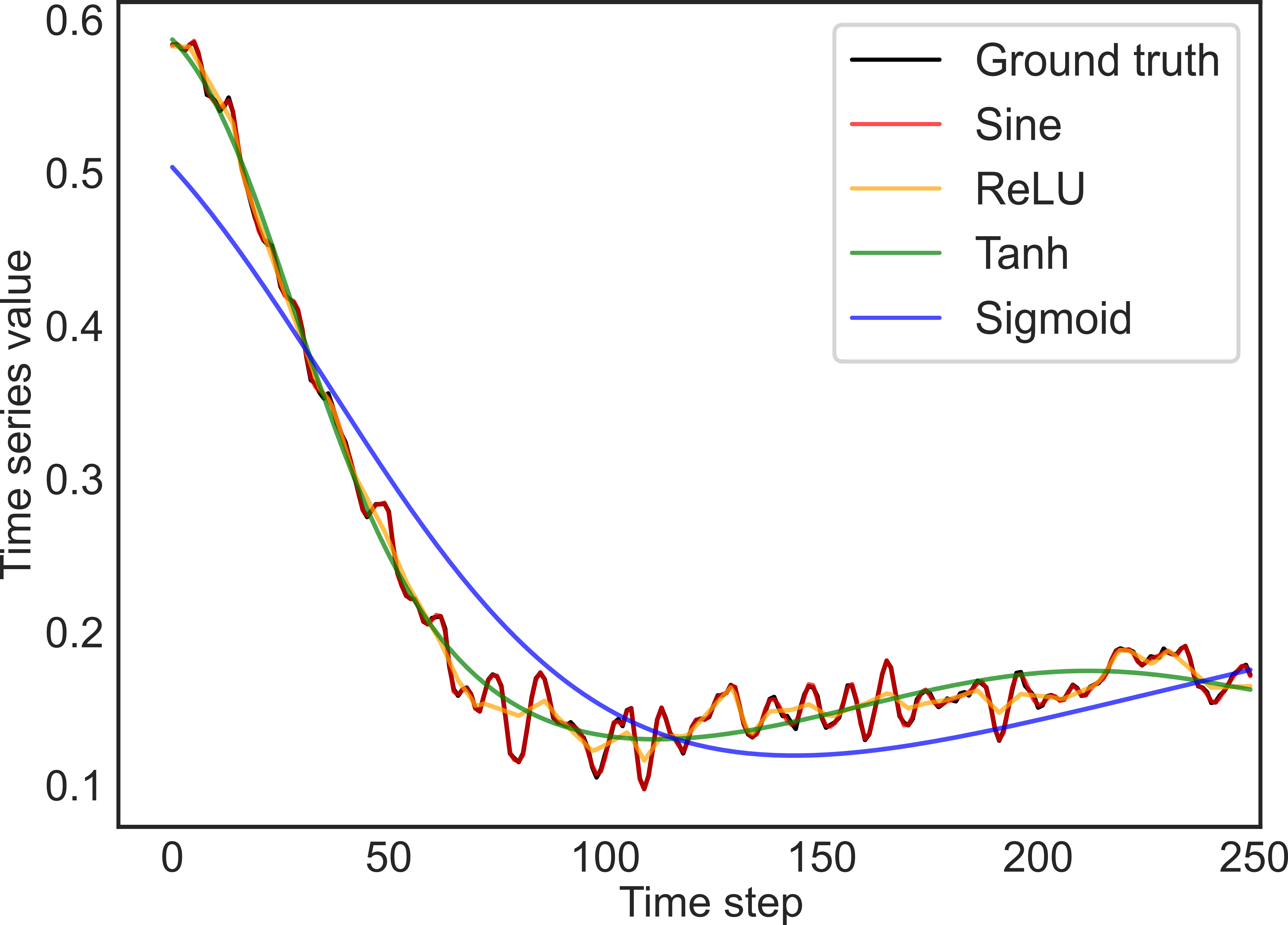}\par\medskip
    {\scriptsize \hspace{4em}Iterations\hspace{22em}Time Steps\par\medskip}
    \caption{Comparison of implicit networks using different activation functions, MSE loss ({\it left}) and reconstructed time-series ({\it right}).}
    \label{fig:recon_uni}
\end{figure}

We start by showing that encoding time series using SIRENS leads to a better reconstruction error than using implicit networks with other activations. We use univariate and multivariate time series datasets from the UCR archive~\cite{UCR2018}.\footnote{The datasets can be downloaded from the project's website: \url{www.timeseriesclassification.com}~\cite{UCRweb}} We selected datasets with different characteristics, either short length time series or long, or in the case of the multivariate datasets, with many features (in some cases, more features than time series length). 
We sample 300 time series (or the maximum number available) from each dataset, train a single SIREN for each time series and calculate the reconstruction error. For comparison we train implicit networks using ReLU, Tanh and Sigmoid activations.
As a sample case, we show in Figure~\ref{fig:recon_uni} the losses (left) and reconstruction plots (right) for one of the univariate datasets (NonInvasiveFetalECGThorax1). Here we observe that sine activations converge much faster, and to lower error values, than other activation functions (for the full set of loss and reconstruction plots, along with a description of each dataset, see additional materials). A summary of results can be found in Table~\ref{tab:recon}, where we observe that the MSE error is at least an order of magnitude lower for sine activations, with respect to other activation layers.

\begin{table}[htb!]
  \caption{Comparison using MSE of implicit networks using different activation functions on different univariate and multivariate time series from the UCR dataset.}
  \label{tab:recon}
  \centering
\begin{tabular}{lcccc}
\toprule
{} &      Sine &      ReLU &      Tanh &   Sigmoid \\
\midrule
\textit{Univariate} & & & & \\
\midrule
Crop                       & {\bf 5.1e-06} & 5.4e-03 & 2.8e-02 & 5.1e-01 \\
NonInvasiveFetalECGThorax1 & {\bf 2.3e-05} & 2.8e-02 & 5.7e-02 & 8.1e-02 \\
PhalangesOutlinesCorrect   & {\bf 7.5e-06} & 1.9e-02 & 1.4e-01 & 3.3e-01 \\
FordA                      & {\bf 9.2e-06} & 1.4e-01 & 1.5e-01 & 1.5e-01 \\
\midrule
\textit{Multivariate} & & & & \\
\midrule
Cricket        & {\bf 1.6e-04} & 4.2e-03 & 5.1e-03 & 1.6e-02 \\
DuckDuckGeese  & {\bf 9.1e-05} & 8.0e-04 & 8.7e-04 & 9.1e-04 \\
MotorImagery   & {\bf 1.7e-03} & 1.1e-02 & 1.1e-02 & 1.8e-02 \\
PhonemeSpectra & {\bf 1.1e-06} & 6.0e-03 & 1.6e-02 & 1.8e-02 \\
\bottomrule
\end{tabular}
\end{table}

\subsection{Imputation}
\label{sec:exp:imputation}

Real world time series data tend to suffer from missing values, in some cases achieving missing rates of up to $90\%$, making the data difficult to use \cite{Fang2020survey}. Motivated by this, we first use a SIREN trained with the available data points to infer the missing values. Besides fitting a SIREN to the time series, we include a simple regularization term based on a total variation (TV) prior~\cite{lysaker2006}. The TV prior samples random points $t_i \in [-1,1]$ from the available points of the time series (those that do not have missing values) and the regularization consists of the L1 norm on the gradient
\begin{equation*}
    p_{TV} = \frac{1}{N} \sum_{i=1}^N |\nabla \Phi_\theta(t_i)|.
\end{equation*}
We compare this approach with common time series imputation methods, {\it mean}, that fills missing values with the average, {\it kNN}, {\it cubic spline} and {\it linear}. We are not solely interested in the reconstruction error, but also in the accurate and smooth reproduction of the Fourier spectrum composition of the original time series. Therefore, we evaluate both the MSE of the time series reconstruction, and the MAE between the Fourier spectra of ground truth and reconstructed, that we name Fourier Error (FFTE).

\begin{figure}[htb!]
    {\scriptsize \hspace{7.5em}NonInv - Linear\hspace{12em}NonInv - SIREN\hspace{12em}NonInv - SIREN + TV\par\medskip}
    {\tiny \raisebox{6em}{\rotatebox[origin=c]{90}{Time series value}}}~\includegraphics[width=0.32\textwidth, trim={0 0 0 37px},clip]{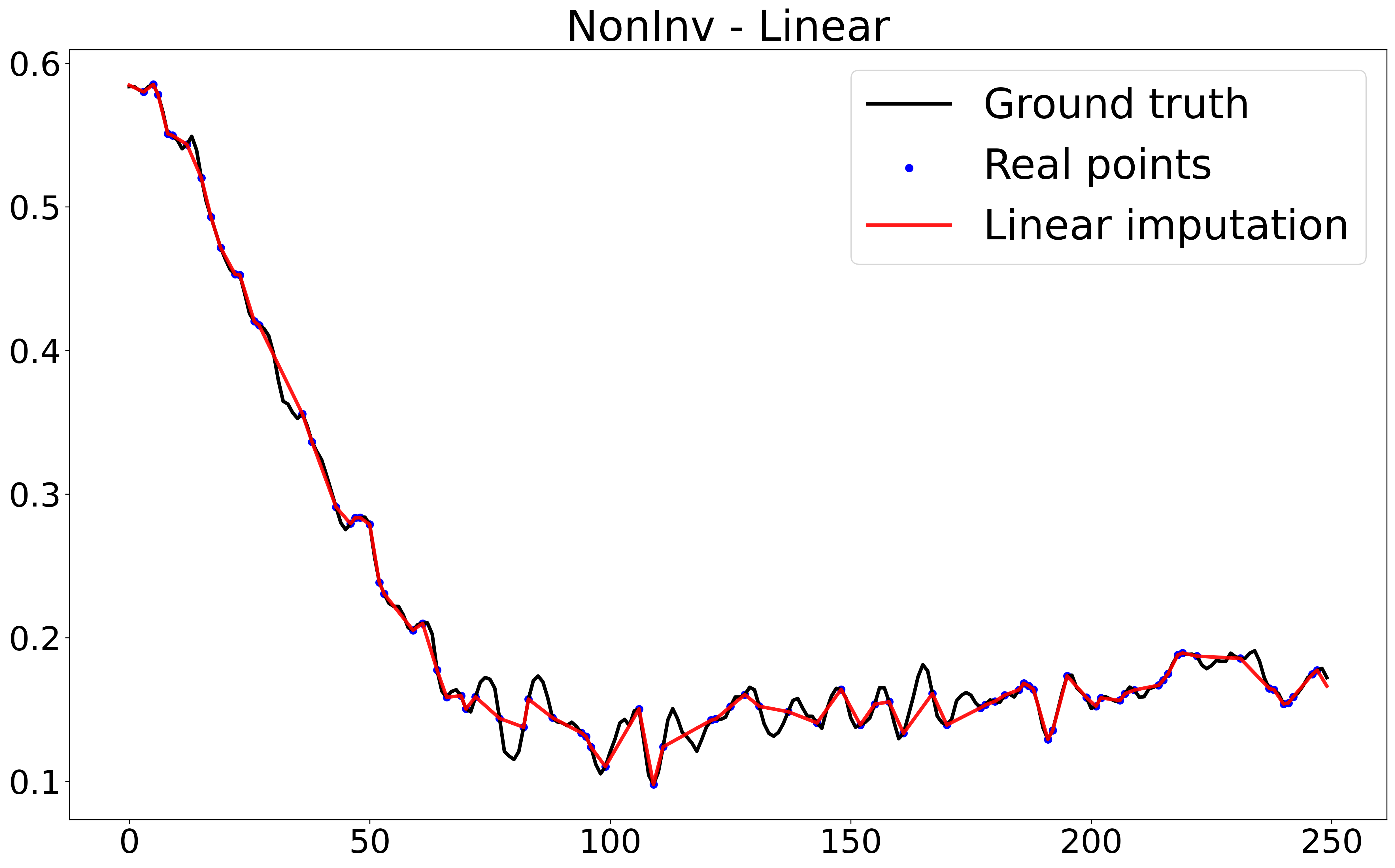}~\includegraphics[width=0.32\textwidth, trim={0 0 0 37px},clip]{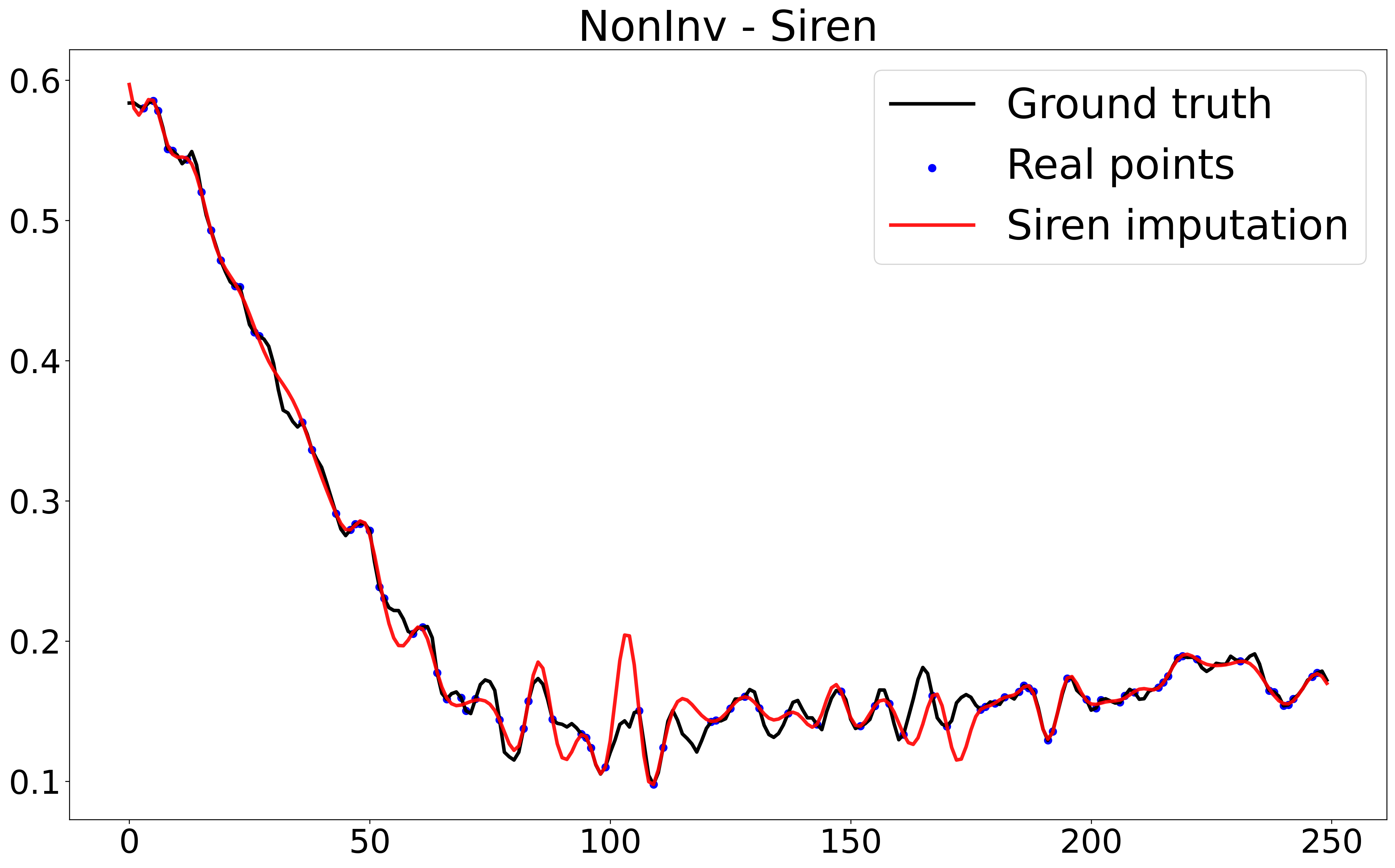}~\includegraphics[width=0.32\textwidth, trim={0 0 0 37px},clip]{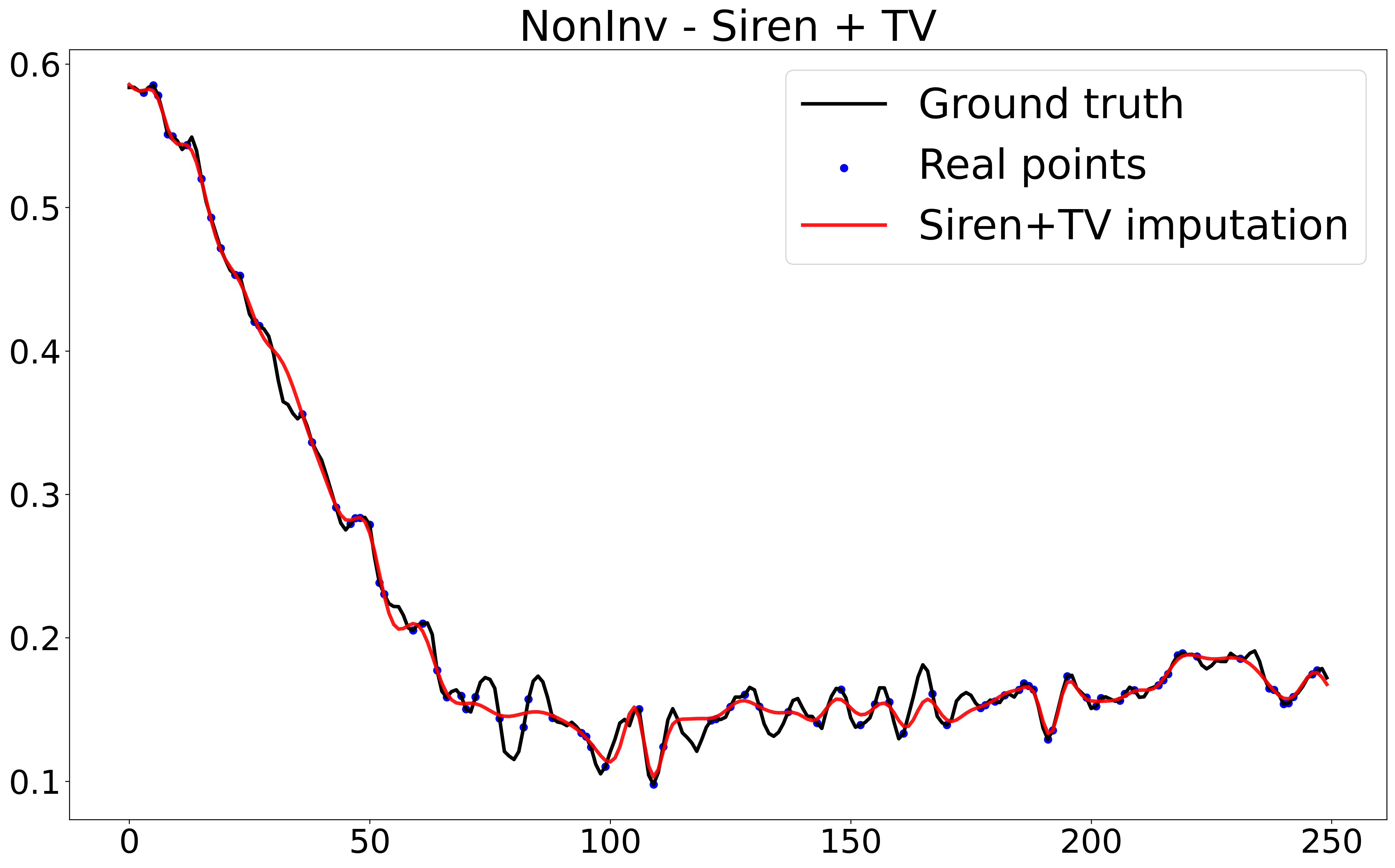}\par\medskip
    {\scriptsize \hspace{8.5em}Time step\hspace{15em}Time step\hspace{14.5em}Time step\par\medskip}
    \caption{Comparison of imputation methods, zoomed in 250 time steps for the NonInv dataset, with $70\%$ of missing values. }
    \label{fig:imputation}
\end{figure}

Table~\ref{tab:imputation_single_siren} shows the comparison of SIREN with and without prior with the classical methods using different ratios of missing values. We can see that the reconstruction error from baseline methods tends to be low, although when we compare the spectrum through FFTE, we can see that the lowest errors are achieved by SIREN, except when we have a fraction of missing values of $90\%$, where in small time series such as Crop or Phalanges, that have less than 100 time steps, both reconstruction error and FFTE are poor. With regard to adding a prior over the gradient, we can see that for very short time series such as Crop and Phalanges (which are 46 and 80 time steps of length, respectively), the prior improves the reconstruction error when compared to SIREN without a prior, and this is also the case for FFTE. Figure~\ref{fig:imputation} shows the comparison of SIREN, SIREN plus TV prior and Linear for a randomly selected time series from the NonInvasiveFetalECGThorax1 (NonInv) dataset with $70\%$ of data points missing (the plot is zoomed in a segment of 250 timesteps to highlight the imputation characteristics). We can see that the linear imputation is not smooth, so it's not a good representation of the series, while SIREN is smooth but tends to have higher deviations from the series. SIREN plus the TV prior is a good compromise between maintaining the smoothness of the series while also deviating less from the ground truth.

\begin{table}[htb!]
    \centering
    \caption{Comparison of implicit neural representations (INRs) with classical imputation methods using different fractions of missing values. Each method shows the MSE reconstruction errors and the Fourier reconstruction errors.}
    \label{tab:imputation_single_siren}
    \resizebox{\textwidth}{!}{%
    \begin{tabular}{llrrrrrrrrrrrr}
        \toprule
        {} & {} & \multicolumn{2}{c}{\bf SIREN} & \multicolumn{2}{c}{\bf SIREN\_TV} & \multicolumn{2}{c}{\bf Mean} & \multicolumn{2}{c}{\bf kNN} & \multicolumn{2}{c}{\bf Cubic Spline} & \multicolumn{2}{c}{\bf Linear} \\
        \\
        {} & {} & {\bf MSE} & {\bf FFTE} & {\bf MSE} & {\bf FFTE} & {\bf MSE} & {\bf FFTE} & {\bf MSE} & {\bf FFTE} &    {\bf MSE} & {\bf FFTE} & {\bf MSE} & {\bf FFTE} \\
        \midrule
		\multirow[c]{5}{*}{\bf Crop}
		& \bf 0.0 & \bfseries 5.5e-07 & \bfseries 0.00 & 1.7e-06 & 0.01 & - & - & - & - & - & - & - & - \\
		& \bf 0.1 & 6.5e-03 & 0.43 & 1.2e-03 & \bfseries 0.18 & 1.7e-02 & 3.01 & 1.7e-02 & 3.03 & \bfseries 6.7e-07 & 2.25 & 2.2e-06 & 2.25 \\
		& \bf 0.5 & 1.2e-01 & 1.80 & 1.3e-02 & \bfseries 0.63 & 1.7e-02 & 3.02 & 1.6e-02 & 2.97 & 8.3e-05 & 2.25 & \bfseries 7.6e-05 & 2.25 \\
		& \bf 0.7 & 2.9e-01 & 2.52 & 3.8e-02 & \bfseries 0.95 & 1.7e-02 & 3.03 & 1.7e-02 & 3.03 & 9.5e-03 & 2.25 & \bfseries 8.6e-04 & 2.24 \\
		& \bf 0.9 & 5.4e-01 & 2.74 & 3.5e-01 & \bfseries 2.02 & 1.7e-02 & 3.02 & 1.7e-02 & 3.00 & 1.1e+02 & 2.49 & \bfseries 1.6e-02 & 2.24 \\
		\midrule
		\multirow[c]{5}{*}{\bf NonInv}
		& \bf 0.0 & \bfseries 1.5e-06 & \bfseries 0.02 & 3.5e-06 & 0.04 & - & - & - & - & - & - & - & - \\
		& \bf 0.1 & 2.1e-06 & \bfseries 0.03 & 4.7e-06 & 0.04 & 1.7e-02 & 3.00 & 1.6e-02 & 2.97 & \bfseries 3.5e-07 & 2.25 & 2.2e-06 & 2.25 \\
		& \bf 0.5 & 9.2e-05 & 0.15 & \bfseries 7.6e-05 & \bfseries 0.14 & 1.7e-02 & 3.00 & 1.7e-02 & 3.02 & 1.1e-04 & 2.25 & 1.1e-04 & 2.25 \\
		& \bf 0.7 & 1.2e-03 & 0.46 & 1.3e-03 & \bfseries 0.45 & 1.7e-02 & 2.99 & 1.7e-02 & 2.99 & 1.7e-02 & 2.25 & \bfseries 6.9e-04 & 2.24 \\
		& \bf 0.9 & \bfseries 1.5e-02 & \bfseries 1.40 & 1.6e-02 & 1.55 & 1.6e-02 & 2.98 & 1.6e-02 & 2.97 & 2.8e-01 & 2.27 & 1.8e-02 & 2.24 \\
		\midrule
		\multirow[c]{5}{*}{\bf Phalanges}
		& \bf 0.0 & \bfseries 4.3e-07 & \bfseries 0.00 & 1.7e-06 & 0.01 & - & - & - & - & - & - & - & - \\
		& \bf 0.1 & 9.0e-04 & 0.22 & 7.6e-04 & \bfseries 0.20 & 1.7e-02 & 3.00 & 1.7e-02 & 3.05 & \bfseries 3.4e-07 & 2.25 & 1.7e-06 & 2.25 \\
		& \bf 0.5 & 2.8e-02 & 1.12 & 1.1e-02 & \bfseries 0.74 & 1.7e-02 & 3.00 & 1.7e-02 & 3.04 & \bfseries 3.5e-05 & 2.25 & 1.1e-04 & 2.25 \\
		& \bf 0.7 & 1.1e-01 & 1.94 & 4.3e-02 & \bfseries 1.27 & 1.6e-02 & 2.99 & 1.7e-02 & 2.99 & 2.3e-03 & 2.25 & \bfseries 6.2e-04 & 2.24 \\
		& \bf 0.9 & 3.3e-01 & 2.75 & 2.2e-01 & 2.26 & 1.6e-02 & 2.99 & 1.7e-02 & 2.99 & 7.1e-02 & 2.26 & \bfseries 1.4e-02 & \bfseries 2.24 \\
		\midrule
		\multirow[c]{5}{*}{\bf FordA}
		& \bf 0.0 & \bfseries 2.1e-06 & \bfseries 0.02 & 4.1e-06 & 0.03 & - & - & - & - & - & - & - & - \\
		& \bf 0.1 & 7.2e-06 & \bfseries 0.04 & 3.3e-05 & 0.09 & 1.7e-02 & 3.00 & 1.7e-02 & 3.02 & \bfseries 5.3e-07 & 2.25 & 1.7e-06 & 2.25 \\
		& \bf 0.5 & 2.0e-03 & \bfseries 0.54 & 5.1e-03 & 0.97 & 1.7e-02 & 3.02 & 1.7e-02 & 3.01 & 9.4e-05 & 2.25 & \bfseries 8.4e-05 & 2.25 \\
		& \bf 0.7 & 1.9e-02 & \bfseries 1.57 & 2.9e-02 & 2.06 & 1.6e-02 & 2.96 & 1.6e-02 & 2.97 & 3.4e-03 & 2.25 & \bfseries 6.4e-04 & 2.24 \\
		& \bf 0.9 & 1.3e-01 & 3.52 & 1.5e-01 & 3.78 & 1.7e-02 & 3.01 & 1.7e-02 & 2.99 & 1.2e-01 & 2.26 & \bfseries 1.6e-02 & \bfseries 2.24 \\
		\bottomrule
	\end{tabular}
	}
\end{table}

\subsection{Time Series Generation}
\label{sec:exp:generation}

\begin{figure}[htb!]
    \centering
    \includegraphics[width=\textwidth]{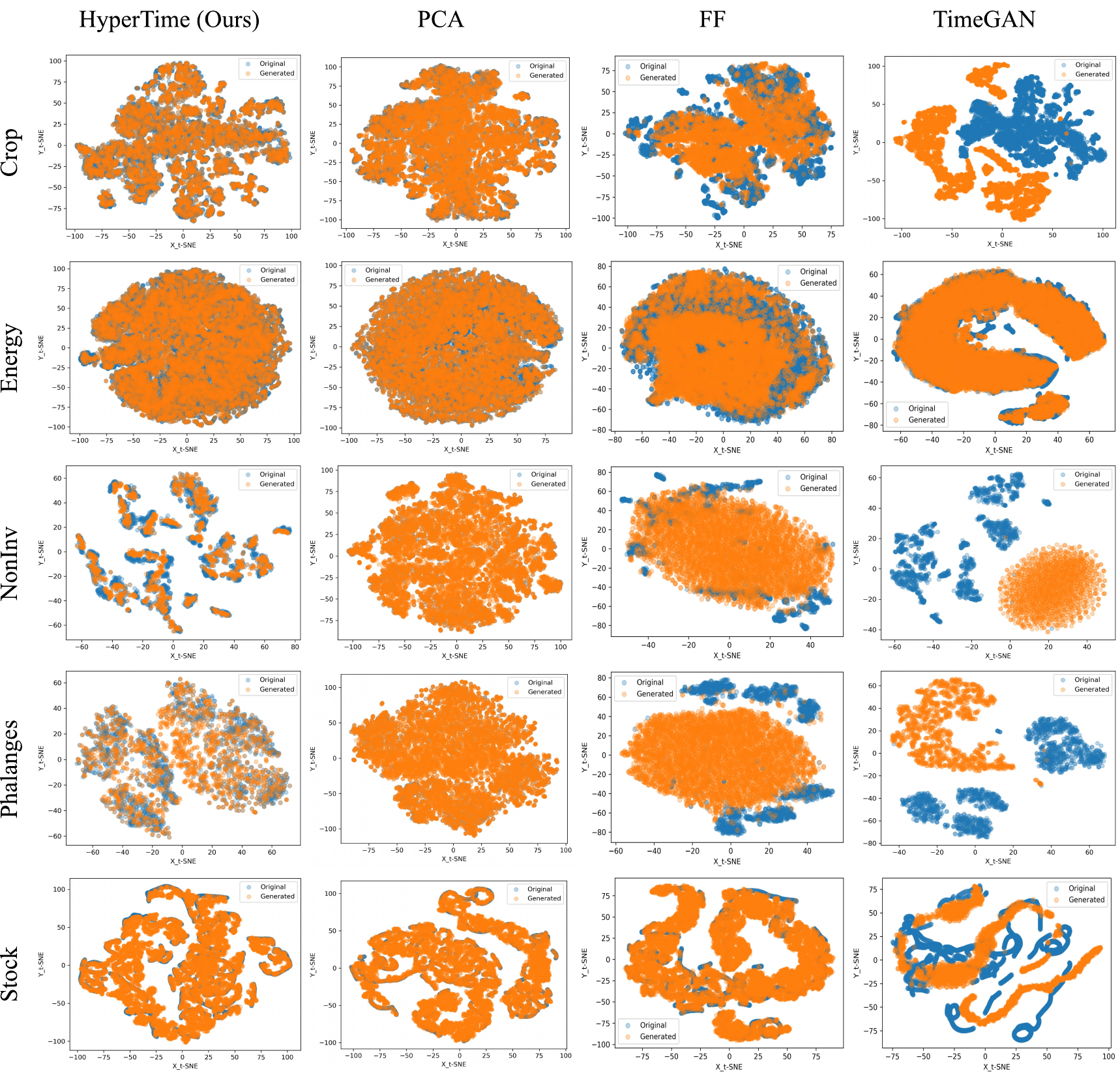}
    \caption{t-SNE visualization on univariate datasets (in rows: Stocks, Energy, Crop, NonInv and Phalanges), using different time series generation methods (in columns: HyperTime, PCA, Fourier Flows and TimeGAN). Blue corresponds to original data and orange to synthetic data.}
    \label{fig:tsne-univariate}
\end{figure}

To evaluate the utility of learning a prior over the space of implicit functions, we use the set encoder network and the hypernetwork to generate new time series. We do so by projecting time series into the latent vector of the HyperTime network and interpolating the latent vector. This is similar to training an autoencoder and interpolating the latent space, but the output of the decoder of HyperTime are the weights of the SIREN networks. We follow the experimental set up proposed in \cite{alaa2021} for the evaluation, were the performance of the synthetic data is evaluated using a predictive score (MAE) that corresponds to the prediction accuracy of an off-the-shelf neural network trained on the synthetic data and tested on the real data. Additionally, to measure the quality of the synthetic data, we use the precision and recall averaged over all time steps, which are then combined into a single F-score. We use the same datasets as before, and we add two datasets that were used in Fourier Flows~\cite{alaa2021} and TimeGAN~\cite{Yoon2019TimeseriesGA}, Google stocks data and UCI Energy data. We compare our HyperTime model with generating data using PCA, with Fourier Flows and TimeGAN, two state-of-the-art methods for time series generation. Table~\ref{tab:gen_scores} shows the performance scores for all models and datasets. Additionally, we visualize the generated samples using t-SNE plots \ref{fig:tsne-univariate} where we can see that the generated data from HyperTime exhibits the same patterns as the original data. In the case of Fourier Flows, in the UCR datasets we see that NonInv and Phalanges do not show a good agreement.

The synthesis of time series via principal component analysis is performed in a similar fashion as our HyperTime generation pipeline. We apply PCA to generate a decomposition of time series into a basis of $40$ principal components. The coefficients of these components constitute a latent representation for each time series of the dataset, and we can interpolate between embeddings of known time series to synthesize new ones. The main limitation of this procedure, besides its linearity, is that it can only be applied to datasets of equally sampled time series.

\begin{table}[htb!]
  \caption{Performance scores for data generated with HyperTime and for all baselines.}

  \label{tab:gen_scores}
  \centering
\begin{tabular}{lccccc} 
\toprule
{}               & \bf Crop  & \bf NonInv & \bf Phalanges & \bf Energy  & \bf Stock \\
\midrule
\bf PCA          & & & & & \\
\it MAE          & 0.050     & 0.019      & 0.050         & \bf 0.007   & 0.110\\
\it F1 Score     & \bf 0.999 & \bf 0.999  & \bf 0.999     & 0.998       & \bf 0.999 \\
\midrule
\bf HyperTime (Ours) & & & & & \\
\it MAE          & \bf 0.040 & \bf 0.005  & \bf 0.026     & 0.058       & 0.013 \\
\it F1 Score     & \bf 0.999 & 0.996      & 0.998         & \bf 0.999   & 0.995 \\
\midrule
\bf TimeGAN      & & & & & \\
\it MAE          & 0.048     & 0.028      &  0.108        & 0.056       & 0.173 \\
\it F1 Score     & 0.831     & 0.914      &  0.960        & 0.479       & 0.938 \\
\midrule
\bf Fourier Flows& & & & & \\
\it MAE          & \bf 0.040 & 0.018      & 0.056         & 0.029       & \bf 0.008\\
\it F1 Score     & 0.991     & 0.990      & 0.992         & 0.945       & 0.992\\

\bottomrule
\end{tabular}
\end{table}

\begin{figure}[htb!]
    \centering
    \includegraphics[width=0.51\textwidth]{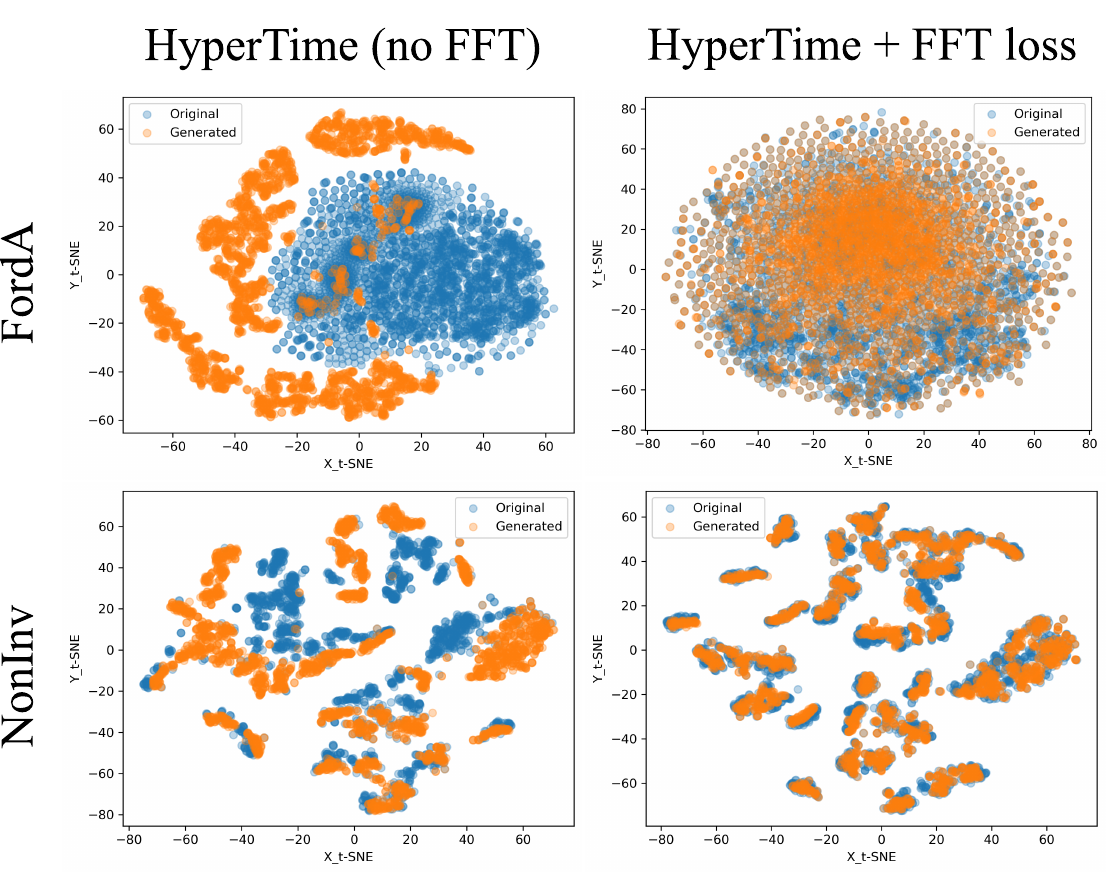}~   \includegraphics[width=0.48\textwidth]{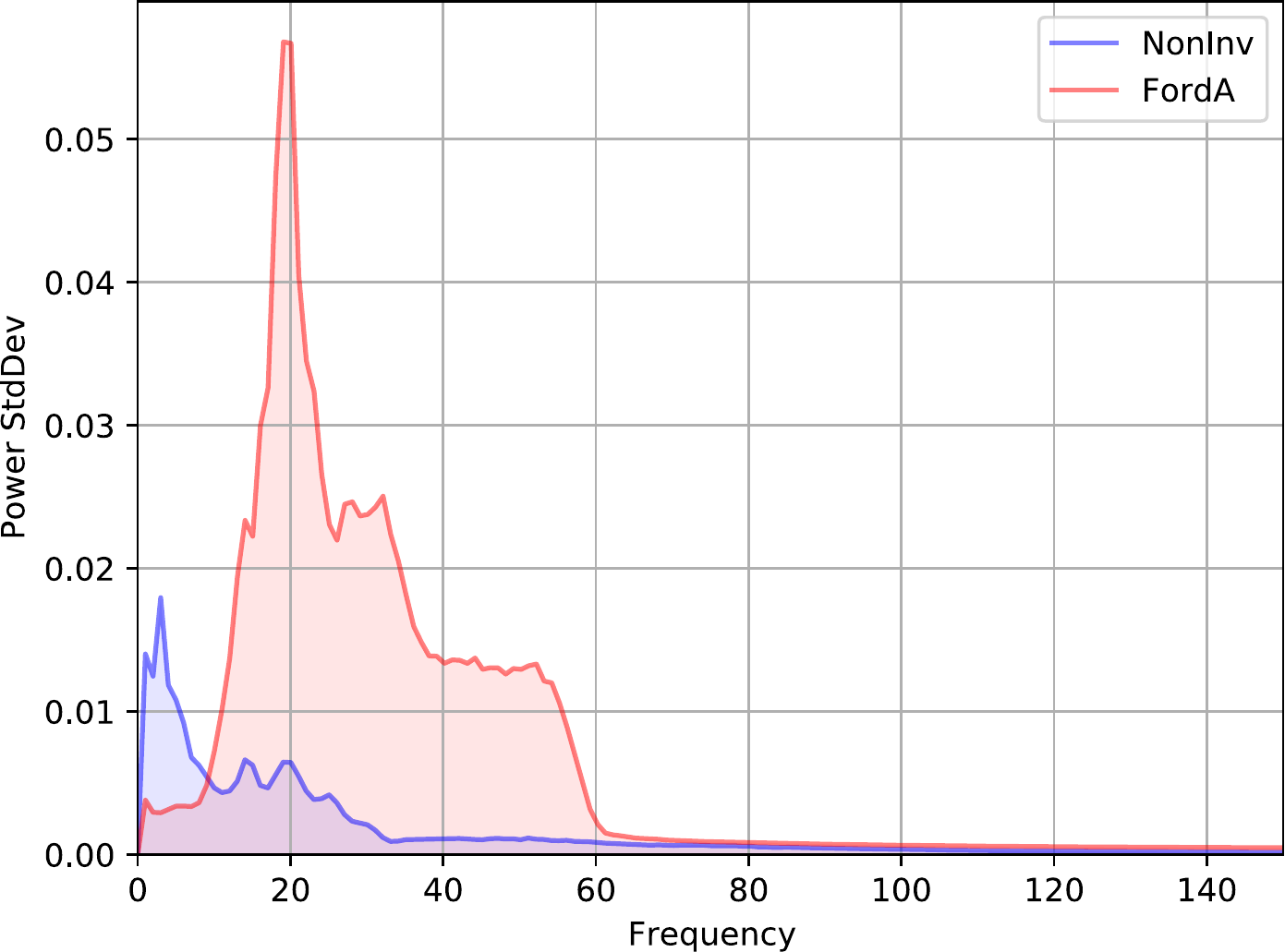}
    \caption{\emph{Left:} t-SNE visualization of ground truth and generated data on two univariate datasets (NonInv and FordA), using HyperNet with and without the Fourier-based loss $\mathcal{L}_\text{FFT}$ (Eq.~\ref{eq:fft_loss}). \emph{Right:} Standard deviation of the power spectra for the time series of the same two datasets. FordA shows a considerably larger number of variations in the distributions of the power spectra, which explains the difficulty of HyperTime to learn patterns from the data.}
    \label{fig:stddev_fft}
\end{figure}

Finally, we analyze the importance of the Fourier-based loss $\mathcal{L}_\text{FFT}$ from equation~\ref{eq:fft_loss} on the training of HyperTime. In Figure~\ref{fig:stddev_fft}-left we display t-SNE visualizations of time series synthesized by HyperTime with and without the use of the FFT loss during training, for two datasets (NonInv and FordA). In both cases, the addition of the $\mathcal{L}_\text{FFT}$ loss results in an improved matching between ground truth and generated data. However, in the case of FordA, the addition of this loss becomes crucial to guide the learning process. This is also reflected in the numerical evaluations from Table~\ref{tab:ablation}, which shows steep improvements in performance for the FordA dataset.

A likely explanation for the difficulty of the network to learn meaningful patterns from the data of this dataset is provided by the right plot in Figure~\ref{fig:stddev_fft}. Here we show the standard deviation of the power spectrum for both datasets, as a function of the frequency. The difference in the distributions indicates that FordA is composed of spectra that present larger variability, while NonInv's spectra are considerably more clustered. 
Further research on the characteristics of the datasets that benefit the more from the $\mathcal{L}_\text{FFT}$ loss should be further investigated, especially focusing on non-stationary time series. 

\begin{table}[htb!]
  \caption{Performance scores for data generated with HyperTime, with and without the Fourier-based loss $\mathcal{L}_\text{FFT}$, for two datasets (NonInvasiveFetalECGThorax1, FordA).}
  \label{tab:ablation}
  \centering
    \begin{tabular}{lcc}
    \toprule
    {}                             & \bf NonInv  & \bf FordA \\
    \midrule
    \bf HyperTime + FFT loss       & & \\
    \it MAE                        & \bf 0.0053    & \bf 0.0076 \\
    \it F1 Score                   & \bf 0.9962      & \bf 0.9987 \\
    \midrule
    \bf HyperTime (no FFT)         & &\\
    \it MAE                        & 0.0058    & 0.1647 \\
    \it F1 Score                   & 0.9960     & 0.0167 \\
    \bottomrule
\end{tabular}
\end{table}


\section{Conclusions}
\label{sec:conc}

In this paper we explored the use of implicit neural representations for the encoding and analysis of both univariate and multivariate time series data. We compared multiple activation functions, and showed that periodic activation layers outperform traditional activations in terms of reconstruction accuracy and training speed. Additionally, we showed that INRs can be leveraged for data imputation, resulting in good reconstructions of both the original data and its power spectrum, when compared with classical imputation methods. Finally, we presented HyperTime, a hypernetwork architecture to generate synthetic data which enforces not only learning an accurate reconstruction over the learned space of time series, but also preserving the shapes of the power distributions. For this purpose, we introduced a new Fourier-based loss on the training, and showed that for some datasets it becomes crucial to enable the network to find meaningful patterns in the data. We leveraged the latent representations learned by the hypernetwork for the generation of new time-series data, and compared favorably against current state-of-the-art methods for time-series augmentation. Besides reconstruction, imputation and synthesis, we believe that both INRs and HyperTime open the door for a large number of potential applications in time-series, such as upsampling, forecasting and privacy preservation.

\paragraph{Disclaimer}
This paper was prepared for informational purposes in part by
the Artificial Intelligence Research group of JPMorgan Chase \& Co\. and its affiliates (``JP Morgan''),
and is not a product of the Research Department of JP Morgan.
JP Morgan makes no representation and warranty whatsoever and disclaims all liability,
for the completeness, accuracy or reliability of the information contained herein.
This document is not intended as investment research or investment advice, or a recommendation,
offer or solicitation for the purchase or sale of any security, financial instrument, financial product or service,
or to be used in any way for evaluating the merits of participating in any transaction,
and shall not constitute a solicitation under any jurisdiction or to any person,
if such solicitation under such jurisdiction or to such person would be unlawful.

\bibliographystyle{plain} 
\bibliography{main}

\end{document}